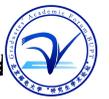

# 基于改进 U-NET 网络的神经分割方法研究

摘要：局部麻醉技术作为现代社会最为常见的麻醉技术，具有安全性高，副作用小等优势。通过分析超声图像，分割图像中的神经区域，有助于提升局部麻醉手术的成功率。卷积神经网络作为目前最为高效的图像处理方法之一，具有准确性高，预处理少等优势。通过卷积神经网络来对超声图像中的神经区域进行分割，速度更快，准确性更高。目前已有的图像分割网络结构主要有U-NET[1]，SegNet[2]。U-NET网络训练时间短，训练参数较少，但深度略有不足。SegNet 网络层次较深，训练时间过长，但对训练样本需求较多由于医学样本数量有限，会对模型训练产生一定影响。本文我们将采用一种改进后的 U-NET 网络结构来对超声图像中的神经区域进行分割，改进后的 U-NET 网络结构加入的残差网络(residual network)[3]，并对每一层结果进行规范化(batch normalization)处理[4]。实验结果表明，与传统的U-NET网络结构相比，改进后的U-NET 网络分割效果具有显著提升，训练时间略有增加。同时，将改进后的 U-NET 网络与 SegNet 网络结构进行对比，发现改进后的 U-net 无论从训练速度还是从分割效果均高于 SegNet 网络结构。改进后的 U-net 网络结构在神经识别方面具有很好的应用场景。

关键字：深度学习,神经分割,卷积神经网络,残差网络

## 一．引言

局部麻醉也称部位麻醉（regional anaesthesia），是指在患者在神志清醒的状态下，将麻药应用于身体局部，使机体某一部分的感觉神经传导功能暂时被阻断，运动神经传导保持完好或同时有程度不等的被阻滞状态。这种阻滞应完全可逆，不产生任何组织损害。与传统的全身麻醉相比，局部麻醉的优点在于简便易行、安全、患者清醒、并发症少和对患者生理功能影响。局部麻醉技术目前广泛应用于外科手术中,局部麻醉技术成功的关键在于能够准确确定神经位置，传统的神经位置的确定主要依赖体表解刨标志定位神经，但是这种方法对于存在很大的风险，容易对神经造成损伤。相比较于传统方法，超声图像具有对风险小，成像快等优点[5]-[7]。目前，越来越多的局部麻醉手术通过超声图像来确定神经区域。本篇论文主要研究通过卷积神经网络分割超声图像中的神经区域。

卷积神经网络作为近年来最为高效的图像处理方法之一，具有准确性高，预处理较少等优势。卷积神经网络目前广泛的应用于语音识别以及医学图像处理领域。将卷积神经网络应用到超声图像分割中[8]-[15]，有助于提高分割神经区域的准确性。

目前已经出现的用于图像分割的网络结构主要有U-NET[1], SegNet[2]，残差网络[3], VGG[16]等。这些网络在图像分割方面都具有显著效果。本文中我们将采用一种改进后的 U-NET 网络对超声图像进行分割。

U-NET 是 Olaf Ronneberger, Philipp Fischer, and Thomas Brox 在 2015 年提出的网络结构[1]，该网络获得了 International Symposium on Biomedical Imaging (ISBI) 2015 竞赛的冠军。U-NET 构造了一个收缩网络和一个扩张网络，形成了一个 U 型结构。收缩过程通过不断的卷积(convolution)和池化(pooling)操作提取图片特征。扩张过程与收缩过程相对应，通过上采样(upsampling)和卷积操作来获取图片的特征。U-NET 的特点在于收缩网络和扩张网络是相互映射的关系，在扩张的过程中，通过合并与之映射的收缩层特征补全丢失的边界信息，提升预测边缘信息的准确性。与SegNet，VGG 等网络结构相比，U-NET 具有训练时间短，结构简单，样本需求少等优势，但是其深度略有不足。

SegNet 是 Vijay Badrinarayanan, Alex Kendall, Roberto Cipolla 在 2016 年提出的卷机神经网络结构[2]，该网络结合了 U-NET，VGG 以及反卷积网络 (deconvolution)[17]中的特点。该网络分为编码层(encoding)与解码层(decoding)两部分，在编码层中，通过不断的卷积操作和池化操作对图像提取特征信息，编码层的网络结构与VGG16网络结构相似。解码层通过不断上采样和卷积操作来进一步提取特征信息。不同于U-NET网络，SegNet 并没有通过合并特征来对边界信息进行补全，SegNet 采用了一种编码器的方法，在编码层中记录下特征的位置信息，并在上采样的过程中，将特征信息映射到相对应的位置，然后进行卷积操作。与 U-NET 相比，SegNet 的网络结构更加深入，层数更多，需要训练样本更多，训练参数更多，训练时间更长，准确性更高。

本文提出了一种改进的 U-NET 结构，该网络结构中引入了残差网以及规范化处理 [7]。残差网络是 Kaiming He，Xiangyu Zhang 于 2015 年提出的网络结构[1]，该结构获得了 ImageNet Large Scale Visual Recognition Competition(ILSVR)竞赛的冠军。残差网络提出了一个"捷径"(shortcut)的概念，即跳过一个或多个层，将输入结果直接添加到底层，残差网络可以通过公式(1)来表示

$$H(x) = x + F(x). \qquad (1)$$

其中$H(x)$为底层的映射，x为输入结果，$F(x)$为网络中的隐藏层输出结果 。

与其他网络相比，残差网络的大大减少了训练参数，增加了网络深度，减少了训练时间。将 U-NET 与残差网络相结合，加深了学习的深度，有助于提高神经区域分割的准确性。实验证明，改进后的 U-NET 分割神经区域的分割效果高于 U-NET 以及 SegNet 网络结构，训练时间略高于U-NET网络，远小于SegNet网络，约为 SegNet 网络的 1/5。

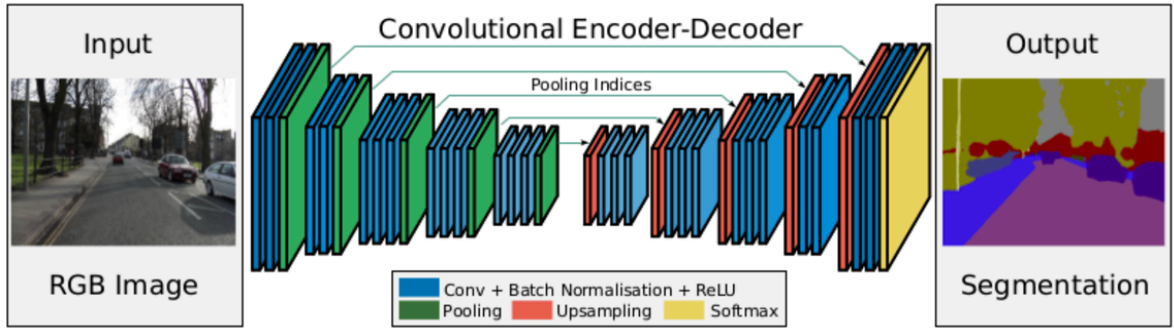

**图 1**. SegNet 网络结构，该结构有编码层与解码层构成，在编码阶段通过卷积操作提取高维特征信息，并将提取的特征的边界信息存入编码器中，最后通过解码器输出特征图，最后进行分类[2]。

## 二．相关工作

### 2.1 U-NET 网络结构

U-NET 网络结构是 2015 年 ISBI 竞赛中提出的网络结构，该结构通过一个收缩网络以及一个扩张网络，构成了一个 U 型结构，对图片进行特征提取，该网络获得了 2015 年 ISBI 竞赛的冠军[1]。

U-NET 网络一共由 23 个卷积层构成，其结构如图 2 所示，收缩网络主要负责下采样的工作，提取高维特征信息，每一次下采样包含两个的3x3的卷积操作，一个 2x2 的池化操作，通过修正线性单元(rectified linear unit, ReLU)作为激活函数，每一次下采样，图片大小变为原来的1/2，特征数量变为原来的 2 倍。扩张网络主要负责上采样的工作，每一次上采样包含两个 3x3 的卷积操作，通过修正线性单元作为激活函数。每一次上采样，图片大小变为原来的 2 倍，特征数量变为原来的 1/2。 在上采样操作中，将每一次的输出特征与相映射的收缩网络的特征合并在一起，补全中间丢失的边界信息。最后，加入 1x1 的卷积操作将之前所获的的特征映射到所属分类上面。

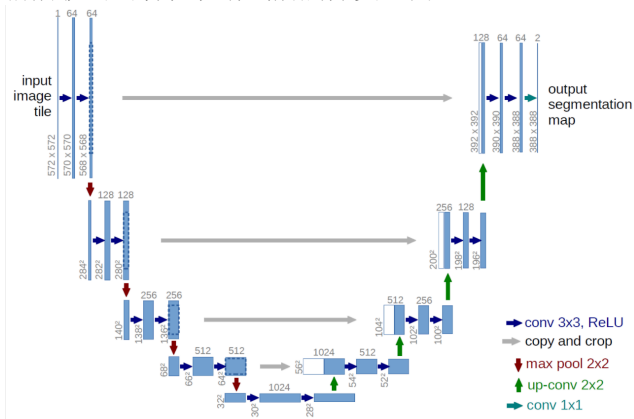

**图 2**.U-NET网络结构，U-NET通过收缩网络和扩张网络构成一个 U 型结构，收缩网络和扩张网络相互映射，白色部分为上采样过程中补充的边界信息。最后通过 1x1 的卷积层输出特征图，进行相应分类[1]。

与其他网络相比，U-NET 具有结构简单，训练时间短，训练参数少等优点，但是与 VGG，SegNet 等网络相比，U-NET 的深度略显不足。

### 2.2 SegNet 网络

SegNet网络是 Vijay Badrinarayanan, Alex Kendall, Roberto Cipolla 于 2016 年提出的神经网络结构[2]，该网络结合了已有的 U-NET 网络，VGG 网络以及反卷积网络(deconvolution network)[15]。该网络在图像分割方面具有显著效果。

SegNet 包含编码层(encoding)和解码层(decoding)两部分，SegNet 网络结构如图 1 所示，编码层的网络结构与 U-NET 的收缩网络类似，但与之不同的是，编码层采用了类似VGG16的网络结构，层数更多，网络更深，并且 SegNet 在每一次卷积之后都会对结果进行规范化处理。解码层类似于 U-NET 中的扩张网络，与编码层相对应，但与 U-NET 不同的是，SegNet 并没有通过合并特征的方法来对边界信息进行补全，SegNet 在编码层进行编码时，通过编码器记录了特征值的边界信息，然后在解码时提取所记录的边界信息。在上采样的过程中，将编码器中的边界信息提取出来，将特征还原到原来位置，然后对其周围进行扩充。解码层在每一次卷积之后对结果进行规范化处理,提高训练速度，提升训练的准确性。最后，通过 1x1 的卷积操作将之前所获的的特征映射到所属分类上面。

与 U-NET 相比，SegNet 的深度更深，效果更好。但是由于其结构较深，训练时间更长，由于医学训练样本相对较少，会对模型训练产生一定影响。

### 2.3 残差网络

在卷积神经网络中，网络层次越深，训练时产生的错误越多，训练时间越长[3]。 残差网络的出现在一定程度上解决了这个问题。残差网络是 Kaiming He, Xiangyu Zhang, Shaoqing Ren, Jian Sun 于 2015 年在 ILSVRC 竞赛中提出的方法，并赢得了 2015 年 ILSVRC 的冠军[3]。

残差网络提出了一种拟合残差映射的方法，即不直接将卷积结果作为输出，而是采用残差映射的方式来进行计算，称之为"捷径"（shortcut）。我们假设

某一隐藏层为F(x)，该层满足映射关系F(x) = H(x) − x，如果多个非线性层组合在一起，我们可以将它们看成一个复杂的网络，我们同样可以假设该隐含层的残差近似于一个复杂的函数，即H(x) = F(x) + x。残差网络的结构如图 3 所示。从图中我们可以看出，残差网络通过将多个卷积层级联的输出与输入相加的方式对图片进行特征提取，减少了训练参数。

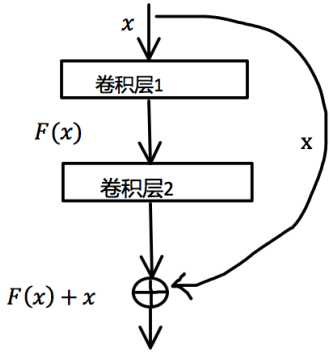

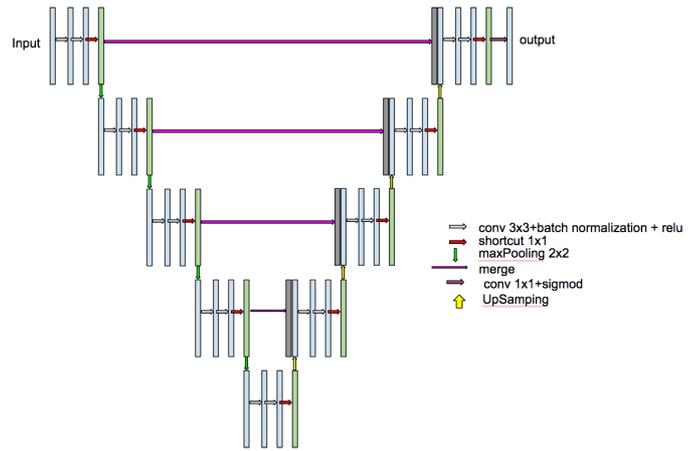

**图 3.** 残差网络结构，残差网络通过将输入结果与多个卷积层级联后的输出结果相加的方式，减少了训练参数，加深网络深度，提高其准确性[3]。

与其他网络相比，残差网络结构相对简单，训练参数较少，训练时间较短，解决了在极深度条件下深度卷积神经网络性能退化的问题，该网络目前广泛应用于计算机视觉中。

## 三． 改进后的 U-NET 模型

由于 U-NET 网络的深度略有不足，SegNet 网络训练参数较多，训练时间较长，本文提出一种新的网络结构，即改进后的 U-NET 网络结构。该结构结合了 U-NET 以及残差网络的特点，将"捷径"的概念引入到 U-NET 网络结构当中，不仅加深了 U-NET 的深度，而且保证了训练时间。改进后的 U-NET 网络与 U-NET 相比，网络结构比 U-NET 更深，训练参数更多，训练时间略有增加，分割效果有显著提升。

### 3.1 改进后的 U-NET 网络

改进后的 U-NET 分为收缩网络以及扩张网络两部分，收缩网络与 U-NET 中的收缩网络类似，有所不同的是，对于每一层输出的结果先进行规范化处理，随后通过激活函数进行激活。每一个上采样包含两个 3x3 的卷积层，一个 1x1 的"捷径"以及一个 2x2 的池化层。每一次下采样都对图片大小变为原来的 1/2，采集的特征数量翻倍。扩张网络与 U-NET 中的扩张网络类似，每一次上采样包含两个 3x3 的卷积层，一个 1x1 的"捷径"，在每一次上采样之前，需要合并收缩网络与之相对应的结果。与收缩网络相似，扩张网络中每一层输出结果都需要先进性规范化处理，随后通过激活函数进行激活。最后，加入 1x1 的卷积网络确定该特征图所对应的结果。改进后的 U-NET 网络结构如图 4 所示。

**图 4.** 改进后的 U-NET 网络结构示意图，与 U-NET 网络相比，改进后的 U-NET 加入了残差网络，并且对于每一层输出结果进行规范化处理。红色的箭头表示"捷径"层，绿色的方块表示通过"捷径"层后获取的结果，灰色方块代表上采样过程中对于边界信息的补充。

加入残差网络后的 U-NET 网络，层次更加深入，训练参数更多，在一定程度上弥补了 U-NET 网络不够深的问题，同时由于残差网络的特性，解决了在极深度条件下深度卷积神经网络性能退化的问题。

### 3.2 "捷径"层

改进后的 U-NET 网络引入"捷径"层（shortcut），"捷径"层的基本网络结构如图 1 所示，本文中我们将该网络结构用公式(2)表示

$$y = W_{n+1}K(w_n x) + bx. \qquad (2)$$

这里 $y$ 和 $x$ 表示该网络的输出以及输入，$W_{n+1}$ 表示权值，K 表示激活函数；b 为一个可调节的参数，本次实验中默认为 1。

一个"捷径"层中可以包含多个卷积层，我们可以将 $W_{n+1}K(w_n x)$ 通过 $F(x,w_i)$ 表示多个卷积层的情况，改进后的公式如(3)所示

$$y = F(x, w_i) + bx. \qquad (3)$$

引入"捷径"层使得 U-NET 的网络结构更加深入，同时也避免了训练时间过长，训练参数过多以及过拟合现象的发生。

### 3.3 损失函数

损失函数(loss function)是用来评估预测值与参考值(ground truth)之间的不一致程度，损失函数越小，模型的鲁棒性越好。本次实验我们将 $L(X,Y)$ 作为该模型的损失函数，$L(X,Y)$ 如公式(4)所示

$$L(X,Y) = \frac{1}{N}\sum_{i=1}^{N} S(X,Y). \qquad (4)$$

(4)

其中 X 表示预测值，Y 表示参考值，$S(X,Y)$ 表示两个模型之间的相似程度，$S(X,Y)$ 表达式如(5)所示。

$$S(X,Y) = 1 - \frac{2|X \cap Y| + k}{|X| + |Y| + k}. \qquad (5)$$

(5)

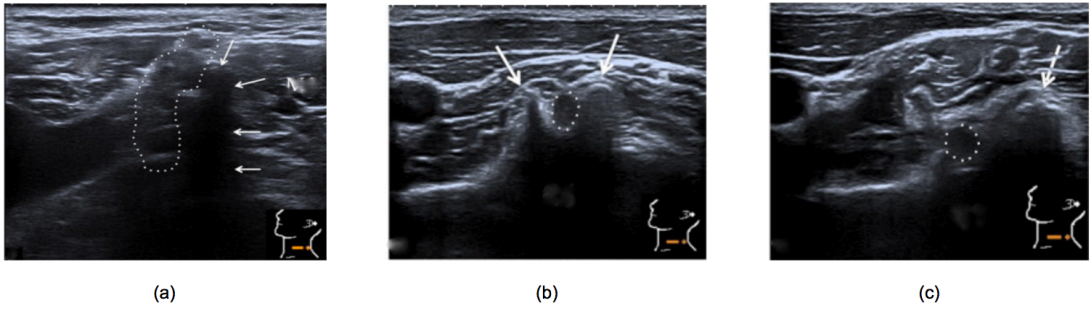

**图 5.** (a),(b),(c)为实验所需的超声图像,每一个图像中圆圈圈出来的为我们需要分割出来的神经部分[19]

**表 1.** 三种网络对应的网络结构

| U-NET | SegNet | 改进后的 U-NET |
|---|---|---|
| $conv3-32$<br>$conv3-32$ | $conv3-64$<br>$conv3-64$ | $conv3-32$<br>$conv3-32$<br>$shortcut-32$ |
| $maxpooling$ | | |
| $conv3-64$<br>$conv3-64$ | $conv3-128$<br>$conv3-128$ | $conv3-64$<br>$conv3-64$<br>$shortcut-64$ |
| $maxpooling$ | | |
| $conv3-128$<br>$conv3-128$ | $conv3-256$<br>$conv3-256$<br>$conv3-256$ | $conv3-128$<br>$conv3-128$<br>$shortcut-128$ |
| $maxpooling$ | | |
| $conv3-256$<br>$conv3-256$ | $conv3-512$<br>$conv3-512$<br>$conv3-512$ | $conv3-256$<br>$conv3-256$<br>$shortcut-256$ |
| $maxpooling$ | | |
| $conv3-512$<br>$conv3-512$ | $conv3-512$<br>$conv3-512$<br>$conv3-512$ | $conv3-512$<br>$conv3-512$ |
| | $maxpooling+upsampooling$ | $shortcut-512$ |
| $upsampooling$+merge | $conv3-512$<br>$conv3-512$<br>$conv3-512$ | $upsampooling$+merge |
| $conv3-256$<br>$conv3-256$ | $upsampooling$ | $conv3-256$<br>$conv3-256$<br>$shortcut-256$ |
| $upsampooling$+merge | $conv3-512$<br>$conv3-512$<br>$conv3-512$ | $upsampooling$+merge |
| $conv3-128$<br>$conv3-128$ | $upsampooling$ | $conv3-128$<br>$conv3-128$<br>$shortcut-128$ |
| $upsampooling$+merge | $conv3-256$<br>$conv3-256$<br>$conv3-256$ | $upsampooling$+merge |
| $conv3-64$<br>$conv3-64$ | $upsampooling$ | $conv3-64$<br>$conv3-64$<br>$shortcut-64$ |
| $upsampooling$+merge | $conv3-128$<br>$conv3-128$ | $upsampooling$+merge |
| $conv3-32$<br>$conv3-32$ | $upsampooling$ | $conv3-32$<br>$conv3-32$<br>$shortcut-32$ |
| | $conv3-64$<br>$conv3-64$ | |
| $conv1-1$ | | |

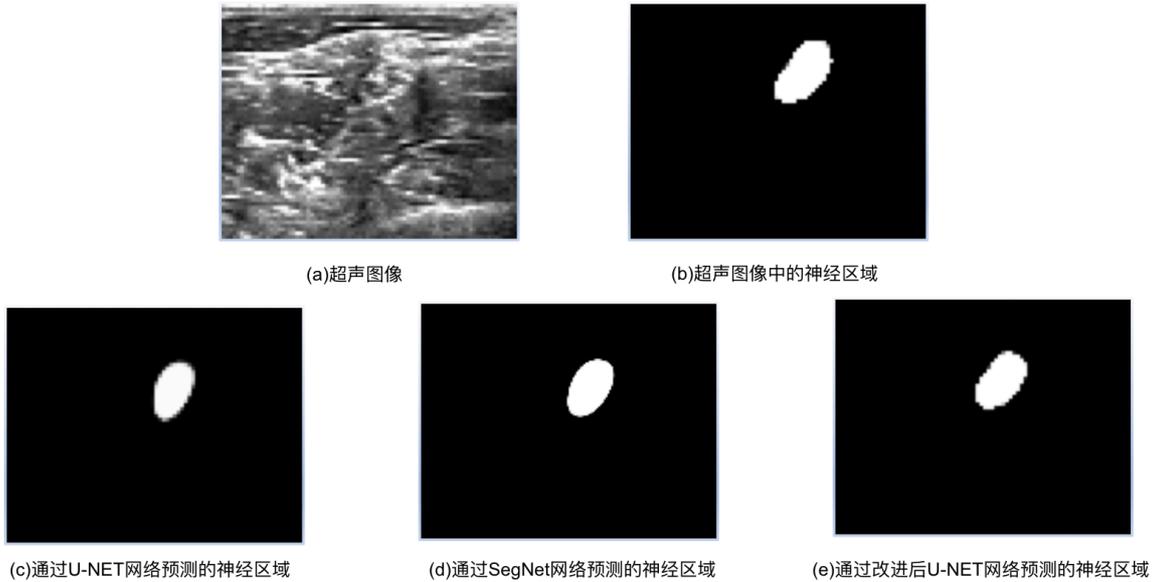

(a)超声图像　　　　　　　　　　(b)超声图像中的神经区域

(c)通过U-NET网络预测的神经区域　　(d)通过SegNet网络预测的神经区域　　(e)通过改进后U-NET网络预测的神经区域

**图 6.** 三种不同网络对同一个超声图像的分割结果，(a)为超声图像，(b)为超声图像中的神经区域，(c)通过 U-NET 网络训练后分割出来的神经区域，(d) 通过 SegNet 网络训练后分割出来的神经区域(e) 通过改进后的 U-NET 网络训练后分割出的神经区域

由于部分超声图像中没有神经区域，所以会出现空图的现象，我们再次加入平滑值 k 对函数进行修正。损失函数介于 0~1 之间，损失函数越小，模型的鲁棒性越好

### 3.4 优化函数

优化函数在训练模型的同时，帮助模型调整权值，使得模型权值调整到最优，使得损失函数最小。本次实验中我们采用 Adam 优化函数，Adam 算法是 Diederik Kingma, Jimmy Ba 于 2014 年提出的优化算法[12]。与其他优化算法相比，Adam 具有计算高效，占用内存较少，善于处理非平稳模型等优势。

## 四．实验及分析

我们采用 kaggle 超声图像神经分割竞赛中的臂神经丛数据分别对 U-NET 网络，SegNet 网络以及改进后的 U-NET 网络进行训练，三种网络的模型如表 1 所示，训练数据大约有 5000 组超声图像。实验结果表明改进后的 U-NET 网络分割效果高于 U-NET 网络和 SegNet 网络。

### 4.1 实验数据

本次实验的实验数据为臂神经丛超声图像。臂神经丛由第五节颈椎神经(C5)到第一节胸椎神经(T1)的前支构成。是头、颈、上肢内连接锁骨、上臂、前臂、手的神经丛的名称。主要支配上肢和肩背、胸部的感觉和运动[18]-[20]。图 5 为部分臂神经丛的超声图像，图中标记出来的部分为我们需要识别出的神经区域。

### 4.2 评价标准

本次实验中我们采用 Dice 系数(dice coefficient)来评价模型的好坏。Dice 系数是一种集合相似度函数，用来评判两个样本之间的相似程度，两个样本相似度越好，Dice 系数越大。Dice 系数如(6)所示

$$s = \frac{2|X \cap Y|}{|X|+|Y|}. \quad (6)$$

其中 X 表示的是预测值，Y 表示的是参考值，$|X \cap Y|$ 表示两个样本之间的相交部分或重叠部分，$|X| + |Y|$ 表示预测值与参考值的总量。本次实验中当预测值与参考值完全相同时，Dice 系数为 1；当预测值与参考值不相关时，Dice 系数为 0。所以 Dice 系数越大，表明两个图像相似度越高，模型越准确。

### 4.3 实验结果以及评估

(1) 实验结果

我们将 kaggle 数据集中的 5000 组图片作为训练集用于训练 U-NET，SegNet 以及改进后的 U-NET 三种网络的网络模型。将三种训练好的网络模型分别对 kaggle 测试集中的数据进行预测，将预测结果提交到 kaggle 网站上进行评估，评估结果如表 2 所示。从表 2 中我们可以看出改进后的 U-NET 网络分割效果明显高于 SegNet 与 U-NET 网络。与 U-NET 网络相比，改进后的 U-NET 网络分割效果提升了 14%，与 SegNet 相比，效果提升了 7%

**表 2** 三种不同网络结构测试结果

| 网络模型 | Dice 系数 |
|---|---|
| U-NET | 0.574 |
| SegNet | 0.611 |
| 改进后的 U-NET | 0.657 |

图 6 为三种不同网络对同一超声图像预测结果的比较。其中(a)图为含有神经区域的超声图像，(b)图为超声图像中的神经区域，(c)(d)(e)分别为采用 U-NET 网络，SegNet 网络以及改进后的 U-NET 网络对(a)图像神经区域的预测图，从图中我们可以看出,改进后的 U-NET 网络分割效果明显好于 U-NET 网络以及 SegNet 网络。

(2) 训练时间评估

表 3 和表 4 分别为三种网络训练参数以及训练时间的比较，表 3 中我们可以看出三种网络训练参数之间的关系，由于改进后的 U-NET 网络引入了残差网络，加深了 U-NET 网络的深度，所以改进后的 U-NET 网络训练参数高于 U-NET 网络，但远小于 SegNet 网络。表 4 为三种网络结构的训练时间的比较，通过引入规范化处理等手段，使的改进后的 U-NET 网络训练时间有所减少。从表 4 中我们可以看出改进后的 U-NET 网络训练时间略高于 U-NET 网络，但远小于 SegNet 网络，训练时间约为 SegNet 网络的 1/5。

表 3. 三种不同网络的训练参数

| 网络模型 | 训练参数 |
| --- | --- |
| U-NET | 7,848,129 |
| SegNet | 31,819,649 |
| 改进后的 U-NET | 8,301,441 |

表 4. 三种不同网络的训练参数

| 网络模型 | 训练时间（min） |
| --- | --- |
| U-NET | 20 |
| SegNet | 343 |
| 改进后的 U-NET | 73 |

(3) 训练过程评估

图 7 为训练过程中为损失函数与迭代次数的关系，通过对比三种模型的训练过程，改进后的 U-NET 网络在训练过程中下降的更多，准确率高于 U-NET 与 SegNet 网络。我们可以得出结论，改进后 U-NET 网络的鲁棒性优于 U-NET 网络与 SegNet 网络，实验结果也证明改进后的 U-NET 网络分割效果高于 U-NET 网络和 SegNet 网络。

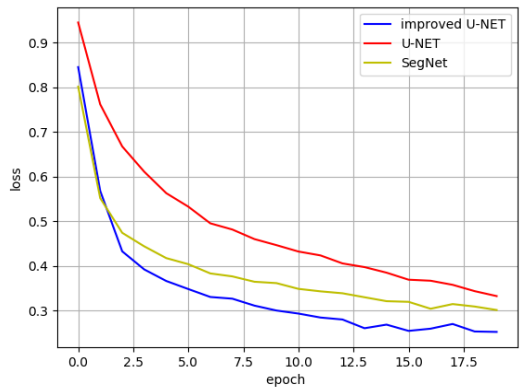

图 7. 网络迭代次数与损失函数的关系，从图中我们可以看改进后的 U-NET 网络结构在训练过程中，要好于 U-NET 与 SegNet 网络。

### 4.4 实验结论

实验表明，改进的 U-NET 网络分割效果优于 U-NET 网络和 SegNet 网络，与 U-NET 网络相比，改进后的 U-NET 网络分割效果提升了 14%，训练时间有所增加；与 SegNet 网络相比，改进后的 U-NET 网络分割效果提升了 7%，训练时间约为 SegNet 的 1/5。实验结果表明，改进后的 U-NET 网络可以更好的分割超声图像中的神经区域。

## 五．总结

为了更好的分割超声图像中的神经区域，提高局部麻醉手术的成功率。本文提出了一种改进后的 U-NET 网络结构。改进后的 U-NET 网络通过引入残差网络增加了 U-NET 网络的深度，使得该网络可以获取图片的高维特征，提升分割图片分割的准确性。通过加入规范化处理，提升了模型的训练速度，增加模型训练的准确性。

实验发现改进后的 U-NET 分割效果较 U-NET 网络与 SegNet 网络结构有很大的提升，训练时间较短，训练参数较少，分割效果较 U-NET 提升了 14%，较 SegNet 提升了 7%。实验结果表明该改进后的 U-NET 结构对于分割神经区域具有显著的效果，可以更加准确的分割超声图像中的神经区域，有助于局部麻醉手术的顺利进行。